\documentclass[10pt,twocolumn,letterpaper]{article}

\usepackage{cvpr}
\usepackage{times}
\usepackage{epsfig}
\usepackage{graphicx}
\usepackage{amsmath}
\usepackage{amssymb}
\usepackage{subfigure}

\usepackage[breaklinks=true,bookmarks=false]{hyperref}

\cvprfinalcopy 

\def\cvprPaperID{6928} 

\ifcvprfinal\fi
\begin{document}

\title{Unit Impulse Response as an Explainer of Redundancy in a Deep Convolutional Neural Network}
\author{Rachana Sathish and Debdoot Sheet\\
Department of Electrical Engineering, Indian Institute of Technology Kharagpur, West Bengal, India\\
{\tt\small rachana.sathish@iitkgp.ac.in, debdoot@ee.iitkgp.ac.in}}

\maketitle

\begin{abstract}
Convolutional neural networks (CNN) are generally designed with a heuristic initialization of network architecture and trained for a certain task. This often leads to over-parametrization after learning and induces redundancy in the information flow paths within the network. This robustness and reliability is at the increased cost of redundant computations. Several methods have been proposed which leverage metrics that quantify the redundancy in each layer. However, layer-wise evaluation in these methods disregards the long-range redundancy which exists across depth on account of the distributed nature of the features learned by the model.  In this paper, we propose (i) a mechanism to empirically demonstrate the robustness in performance of a CNN on account of redundancy across its depth, (ii) a method to identify the systemic redundancy in response of a CNN across depth using the understanding of unit impulse response, we subsequently demonstrate use of these methods to interpret redundancy in few networks as example. These techniques provide better insights into the internal dynamics of a CNN. 
\end{abstract}

\section{Introduction}
\label{sec:intro}
Convolutional Neural Network (CNN) are widely used for various computer vision tasks ranging from image classification, object detection and segmentation to image super-resolution and so on. The architectural complexity of these networks in terms of trainable parameters and number of computations have increased significantly over the past few years owing to the availability of high performance computing resources for offline training. Heuristic design of these models results in redundancies.
Quantitative estimation of redundancy in CNNs have been explored for removing redundant parameters for the purpose of deployment on resource constrained edge devices. These techniques leverage the redundancies in the network to decrease the number of parameters so as to deploy them on edge devices with constrained computational capabilities. Structural pruning of CNNs which removes redundancies at the kernel level uses certain quantitative measures like the summation of values of kernels \cite{su2018redundancy}, threshold \cite{han2015learning}, statistics information \cite{luo2017thinet}. 

Most methods for model redundancy estimation disregard the distributed nature of the features \cite{lecun2015deep} learned by the network. Functionality of individual units in a neural network cannot be comprehended fully since it is also affected by other units along the depth of the network \cite{frosst2017distilling}. Also, qualitative analysis and interpretation is lacking in existing methods. In this work, we explore a set of methods aimed at understanding the source of robustness in a CNN and how its response is affected by altering the composition of different layers. Instead of measuring kernel redundancy based on the features learned by the individual kernel or the activation generated by them while using the training data, we propose to analyze the unit impulse responses.

In this work, we explore a set of methods aimed at understanding the source of robustness in a CNN and how its response is affected by altering the composition of different layers. Further, we also empirically present the ineffectiveness of layer-wise pruning strategies in removing redundancy. Instead of measuring kernel redundancy based on the features learned by the individual kernel or the activation generated by them for the training data, we propose to analyze the unit impulse response of the convolution operations across depth of the network.


\section{Quantifying redundancy in a convolutional neural network}
\label{sec:quant}

\subsection{Theoretical backdrop}
\label{subsec:quant_th}

\begin{figure}[h]
\centering
\includegraphics[width=0.35\textwidth]{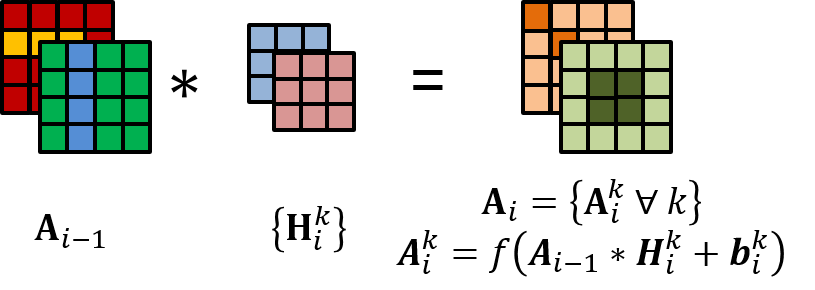}
\caption{In the process of \emph{channel shuffling}, the input $\mathbf A_{i-1}$ refers to the input to the $i^{th}$ layer which convolves with $k$ kernels constituting the learnable weights $\mathbf{H}_i=\{\mathbf{H}_i^k\}$ and results in the output tensor $\mathbf{A}_i$. $f$ denotes the non-linear scalar transformation and $k$ denotes the channel index.}
\label{fig:model}
\end{figure}

\begin{figure}[h]
\centering
\includegraphics[width=0.35\textwidth]{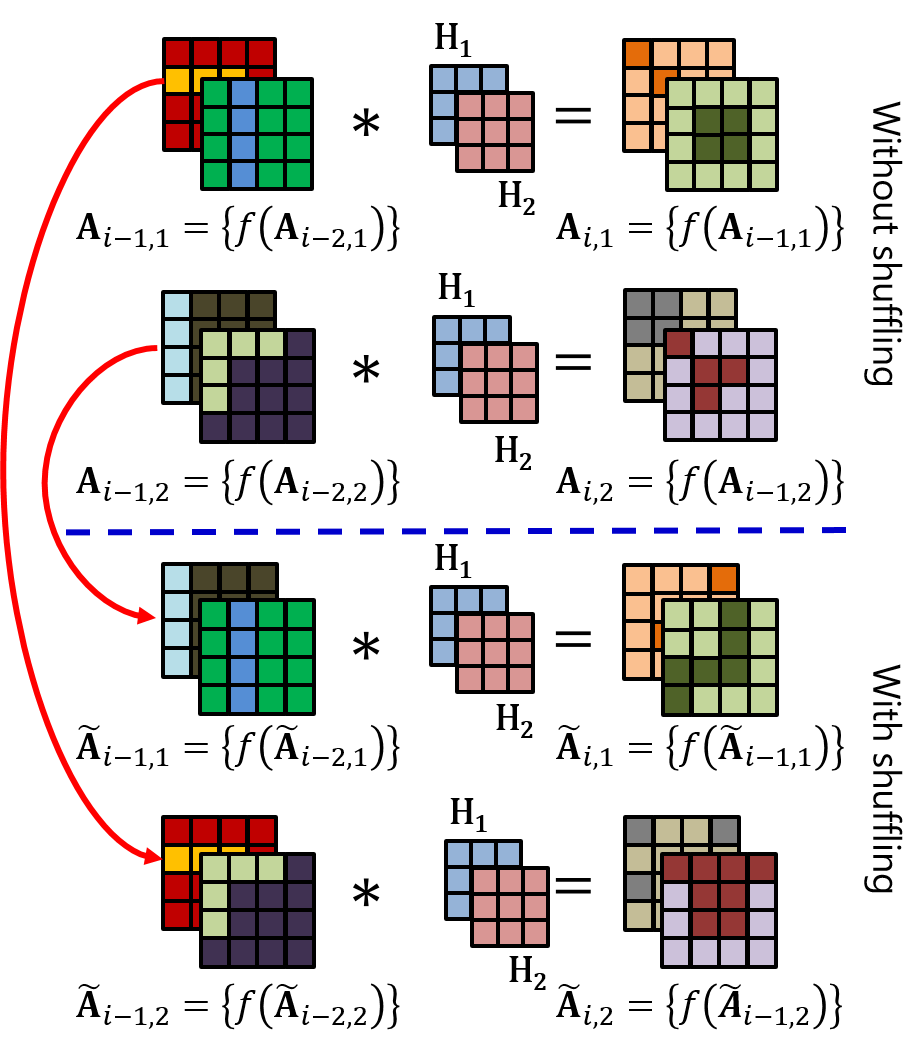}
\caption{Shuffling one channel of input to a convolutional layer, across samples.}
\label{fig:shuffexpt}
\end{figure}

We present methods that will help gain better insight into multifaceted redundancy in a CNN contributing to its robustness. Here, a CNN is analyzed empirically to quantitatively validate existence of redundancy and to understand the causes. Consider two input samples to a CNN, $\mathbf{I}_1$ and $\mathbf{I}_2$ for which $\mathbf{ A}_{i,1}$ and $\mathbf{A}_{i,2}$ are the activation obtained from the $i^{th}$ convolutional layer. Figure~\ref{fig:model} shows a set of kernels in the $i^{th}$ layer and its activation. Robustness of the network is demonstrated by perturbing these activation and observing its effect on the accuracy of prediction. We shuffle the $k^{th}$ channel of the activation $A_{i-1}^{k}$ across all samples to obtain modified activation maps $\mathbf{\tilde A}_{i-1}^k$ as shown in Figure.~\ref{fig:shuffexpt}. We refer to this process as the \emph{`channel shuffling'}.

\begin{figure}[h]
\centering
\includegraphics[width=0.35\textwidth]{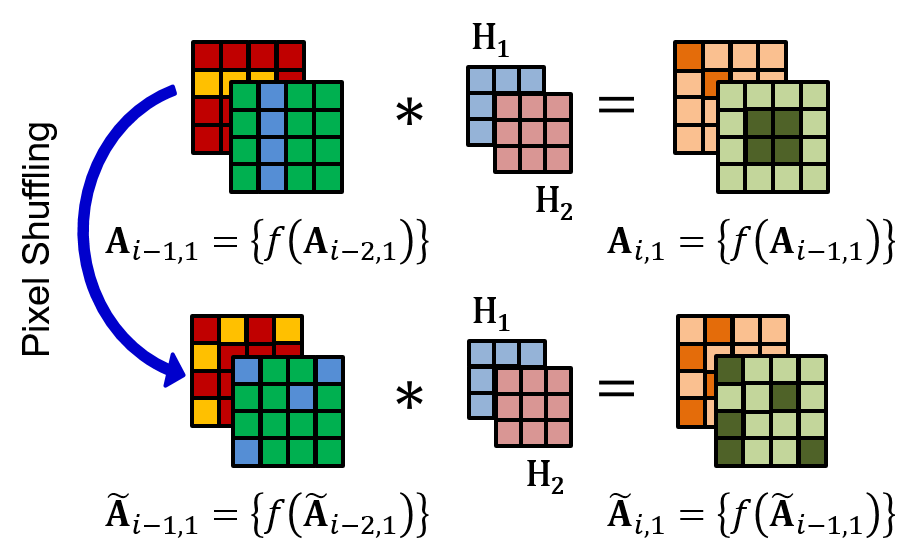}
\caption{Shuffling pixels in one channel of input to a convolutional layer.}
\label{fig:pixelshuff}
\end{figure}

We further explore the robustness of the network by shuffling the pixels within one channel of the input tensor to a layer as shown in Figure~\ref{fig:pixelshuff}. This process is referred to as the \emph{`pixel shuffling'}. 

The change in activation of the $i^{th}$ layer on account of either of the shuffling of the input is measured in terms of cosine similarity (CS) between the corresponding channels of $\mathbf A_i$ and $\mathbf{\tilde{A}}_i$. A robust network is expected to have a minimal change in the activation on account of \emph{channel shuffling} or \emph{pixel shuffling}.

 CS for the $k^{th}$ channel of the activation is given by,
\begin{equation}
CS (\mathbf{A}_{i}^k,\mathbf{\tilde{A}}_{i}^k ) = \frac{\mathbf{A}_{i}^k \cdot \mathbf{\tilde{A}}_{i}^k}{\|\mathbf{A}_{i}^k\|_2\|\mathbf{\tilde{A}}_{i}^k\|_2}
\end{equation}

The effect of shuffling of input can also be measured in terms of change in accuracy of prediction while solving the classification inference on the test dataset.

\subsection{Experiments and Results}
\label{subsec:quant_expt}
\begin{figure}[h]
    \centering
    \includegraphics[width=0.45\textwidth]{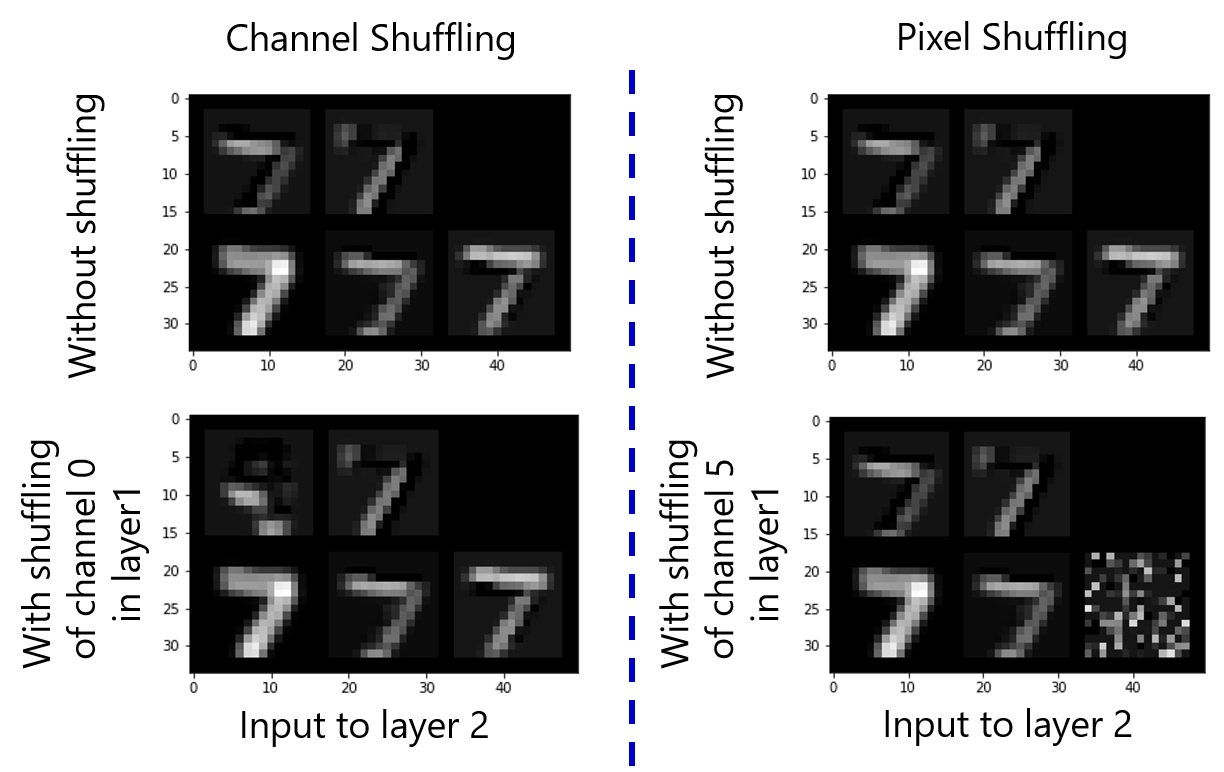}
    \caption{Channel and pixel shuffling of input to the second layer of LeNet-5 for a sample test data from MNIST dataset.}
    \label{fig:shuff_ex}
\end{figure}

\begin{figure}[h]
    \centering
    \subfigure[LeNet-5: Channel 0]{\includegraphics[width=0.23\textwidth]{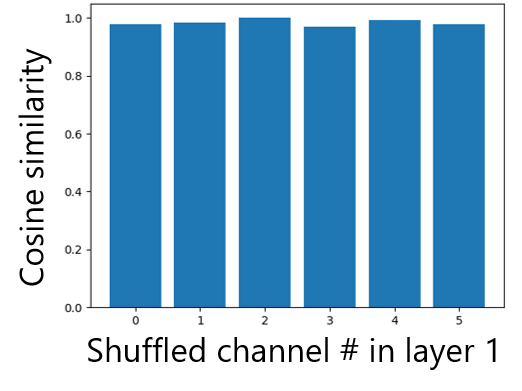}\label{subfig:lenet-chsh-cs-ch0}}
    \subfigure[LeNet-5: Channel 10]{\includegraphics[width=0.23\textwidth]{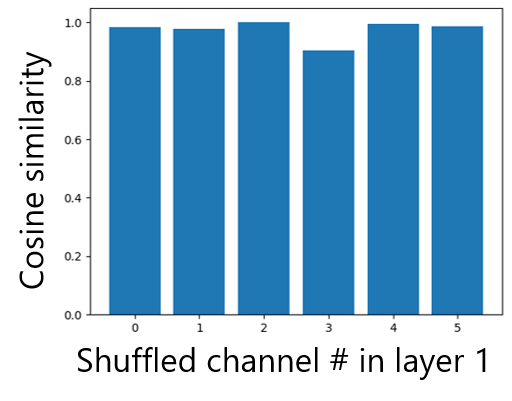}\label{subfig:lenet-pxsh-cs-ch10}}

     \subfigure[LeNet-5x10: Channel 83]{\includegraphics[width=0.23\textwidth]{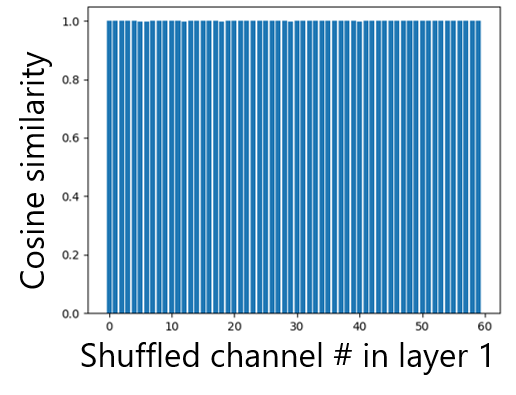}\label{subfig:lenet5x10-chsh-cs-ch83}}
    \subfigure[LeNet-5x10: Channel 92]{\includegraphics[width=0.23\textwidth]{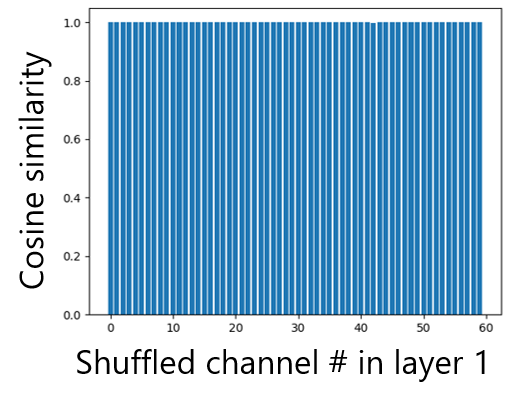}\label{subfig:lenet5x10-pxsh-cs-ch92}}
    \caption{Figure shows the plot of average cosine similarity between the original activation of layer 2 and the ones when one of the input channels chosen at random are perturbed in the (a) \& (c) channel shuffling and (b) \& (d) pixel shuffling experiment.}
    \label{fig:lenet5-chsh-cs}
\end{figure}

\noindent \emph{Channel Shuffling}: To validate the effectiveness of the empirical investigations, we perform the experiment on two CNNs, LeNet-5 \cite{lecun1998gradient} and an over-parametrized version of it, where, the number of learnable parameters in each layer is increased by a factor of $10$ and is referred to as LeNet-5x10. LeNet-5 network having $6$ channels in the first layer and $16$ channels in the second is trained on the MNIST dataset\footnote{http://yann.lecun.com/exdb/mnist/} till convergence. $\mathbf{A}_1^k$ is then shuffled across all samples in each mini-batch of size $256$ in the test set. Figure~\ref{fig:shuff_ex} shows shuffling on a sample data. LeNet-5x10 is also trained till convergence on the MNIST dataset and the experiment repeated. The plot of CS between $\mathbf{A}_2$ and $\mathbf{\tilde{A}}_2$ for each $k$ for LeNet-5 and LeNet-5x10 are shown in Figure~\ref{subfig:lenet-chsh-cs-ch0} and Figure~\ref{subfig:lenet5x10-chsh-cs-ch83} respectively .

\noindent \emph{Pixel Shuffling}:
Pixel shuffling experiment is also performed for each channel $k$ of $\mathbf{A}_1$ of LeNet-5 and LeNet-5x10. Figure~\ref{subfig:lenet-pxsh-cs-ch10} shows the plot of CS for LeNet-5 and Figure~\ref{subfig:lenet5x10-pxsh-cs-ch92} corresponds to LeNet-5x10.

\begin{figure}[h]
    \centering
    \subfigure[LeNet-5: Channel shuffling]{\includegraphics[width=0.23\textwidth]{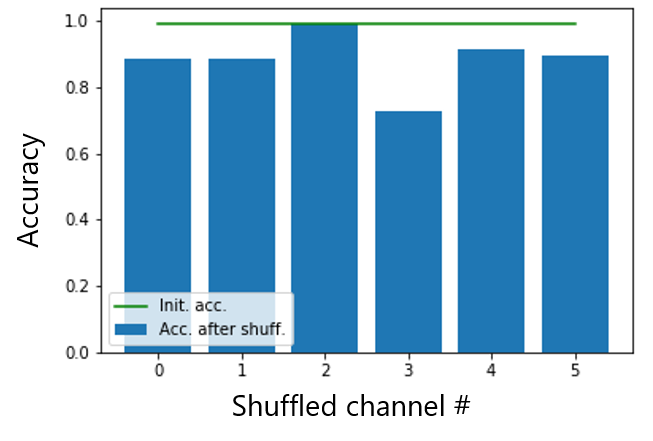}}
    \subfigure[LeNet-5x10: Pixel shuffling]{\includegraphics[width=0.23\textwidth]{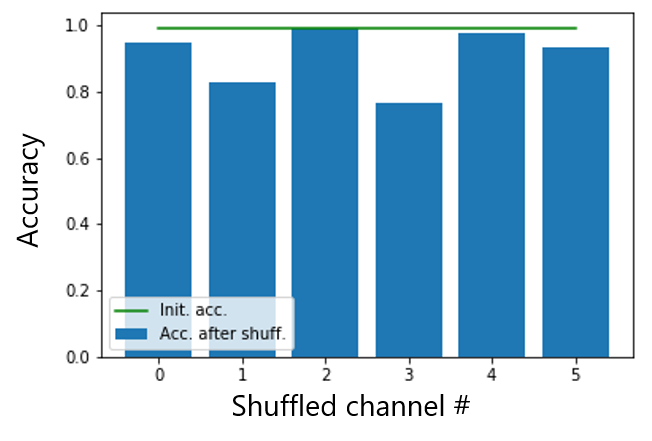}}
    
    \subfigure[LeNet-5: Channel shuffling]{\includegraphics[width=0.23\textwidth]{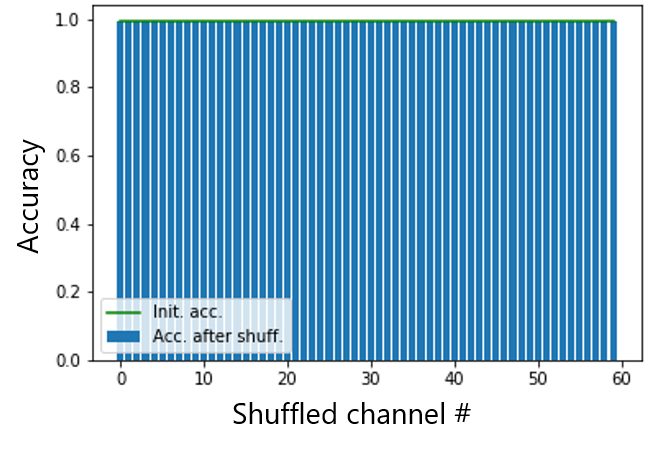}}
    \subfigure[LeNet-5x10:Pixel shuffling]{\includegraphics[width=0.23\textwidth]{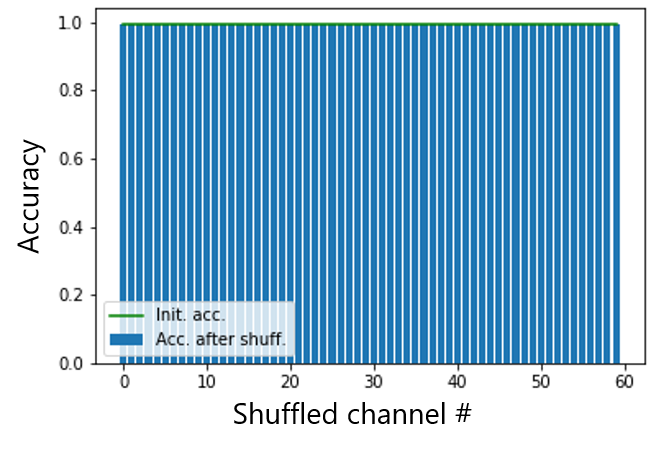}}
    
    \caption{Effect of channel and pixel shuffling on accuracy of LeNet-5 and LeNet-5x10. The horizontal green line denotes the baseline accuracy of the network without any shuffling of the inputs.}
    \label{fig:lenet5-acc}
\end{figure}

   Figure~\ref{fig:lenet5-acc} shows the plots of accuracy of the two network for both the experiments.

\subsection{Discussion}
\label{subsec:quant_disc}
The effect of shuffling of input on the performance of the networks demonstrates the robustness of the CNNs. In the channel and pixel shuffling experiments, if shuffling certain channel $k$ of $\mathbf{A}_1$ does not yield significant change in the cosine similarity of $\mathbf{A}_2$, it can be inferred that the $k^{th}$ channel is redundant. It was observed that, the cosine similarity measured in both the shuffling experiments stayed consistently high for LeNet-5x10 across several channels. Whereas, in the case of LeNet-5, cosine similarity dropped in certain cases.

Further, Figure~\ref{fig:lenet5-acc} shows that shuffling (channel and pixel) had very less impact on the performance of LeNet-5x10 in comparison with LeNet-5, irrespective of the channel being perturbed. This suggests that LeNet5-x10 is more robust. Thus, it can be inferred that the kernels corresponding to the channels which produced the least change in performance measured in terms of accuracy add redundancy to the network. However, these experiments do not answer the question of how redundancy exists in the network.


\section{Unit impulse response as an explainer of redundancy}
\label{sec:impulse}
\subsection{Explaining effect of cascaded convolutions}
\label{subsec:imp_th}

\begin{figure}[h]
    \centering
    \includegraphics[width=0.4\textwidth]{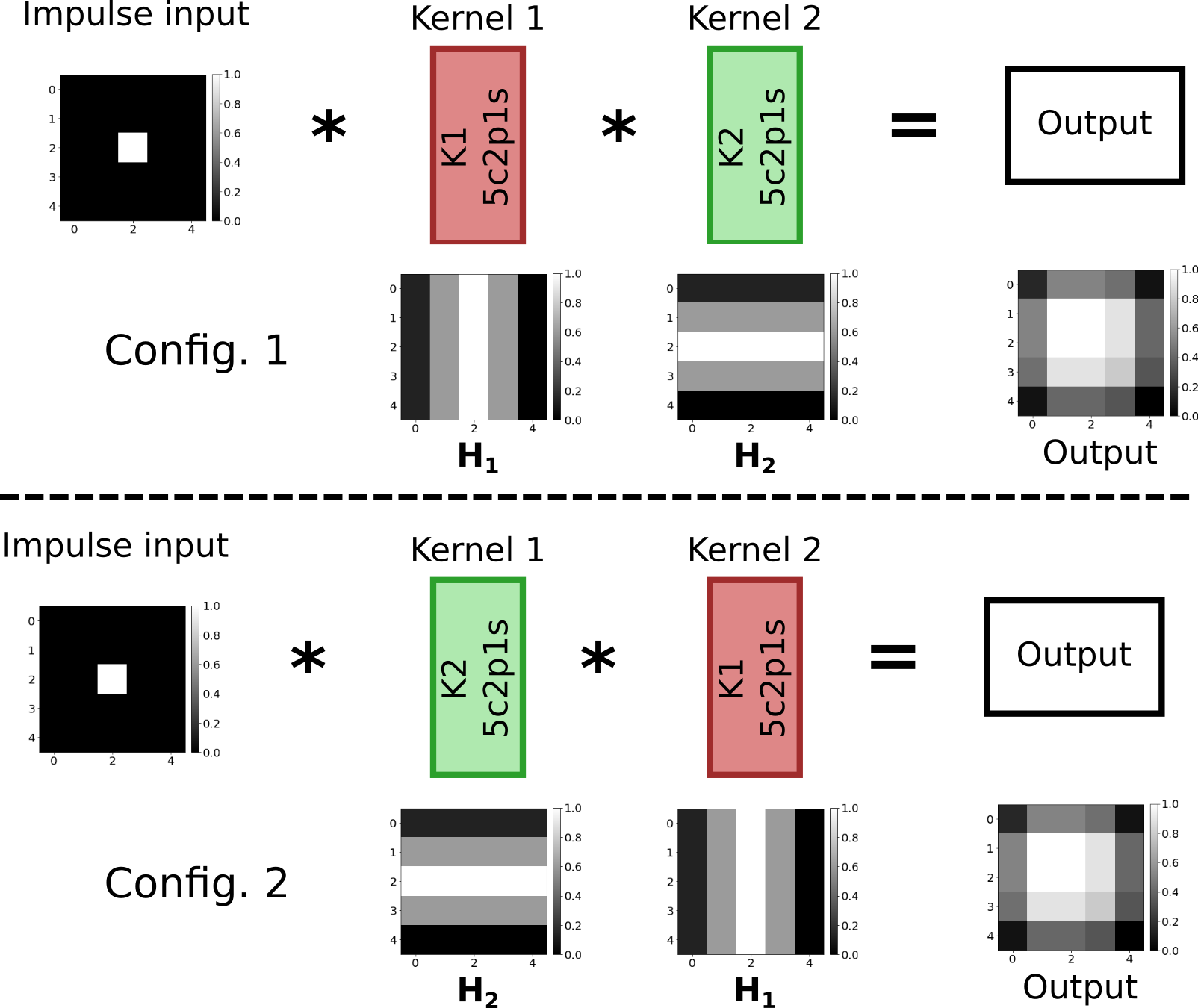}
    \caption{Figure shows the impulse response of a system consisting of two convolutions. The convolutional kernels K1 and K2 are vertical and horizontal Gaussian kernels respectively.}
    \label{fig:casconv}
\end{figure}

In order to explain the the effect of cascaded convolutions as in a CNN, let us  consider the system shown in Figure~\ref{fig:casconv}. It consists of two cascaded convolutions which are equivalent to two layers in a CNN. The position of convolutional kernels $\mathbf{H_1}$ and $\mathbf{H_2}$ are interchanged in configurations 1 and 2. However, the output of the system remains unaffected. In a similar fashion, duplicate kernels may exist along the depth of an over-parametrized CNN. The hierarchical nature of the computations in the network can thus result in similar activation in deeper layers.

\subsection{Theoretical backdrop}
\label{subsec:imp_th}
We propose unit impulse response analysis of a CNN for understanding the redundancies. An impulse input is given as the input to a trained CNN to obtain the impulse response of each convolutional layer. Further, combination of kernels across the depth that result in similar response can be identified by measuring the similarity between the impulse response at each layer measured using normalized cross-correlation.

\subsection{Experiments and Results}
\label{subsec:imp_expt}
\begin{figure}[h]
    \centering
    \subfigure[LeNet-5: Layer 1]{\includegraphics[width=0.15\textwidth]{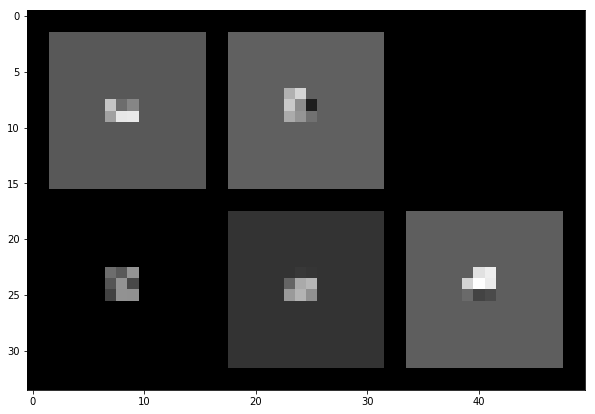}}
    \subfigure[LeNet-5: Layer 2]{\includegraphics[width=0.15\textwidth]{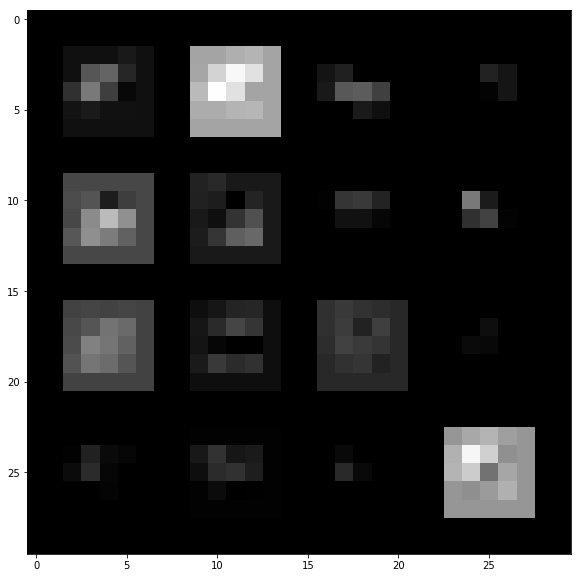}}
    
    \subfigure[LeNet-5x10: Layer 1]{\includegraphics[width=0.17\textwidth]{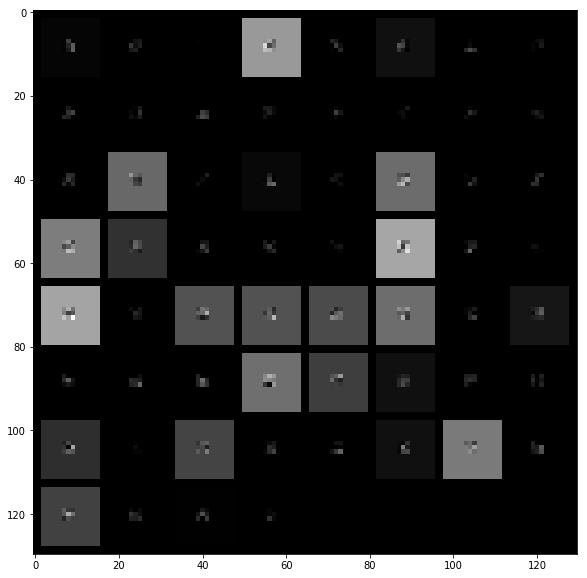}}
    \subfigure[LeNet-5x10: Layer 2]{\includegraphics[width=0.17\textwidth]{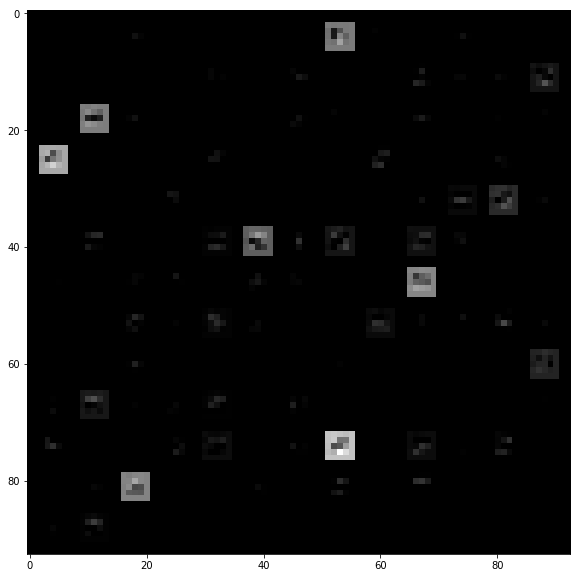}}
    \caption{Figure shows the impulse response of kernels in (a) first and (b) second layer of LeNet-5; (c) first and (d) second layer of LeNet-5x10 respectively. }
    \label{fig:lenet_impresp}
\end{figure}

\begin{figure}[h]
    \centering
    \subfigure[LeNet-5: Layer 1]{\includegraphics[width=0.14\textwidth]{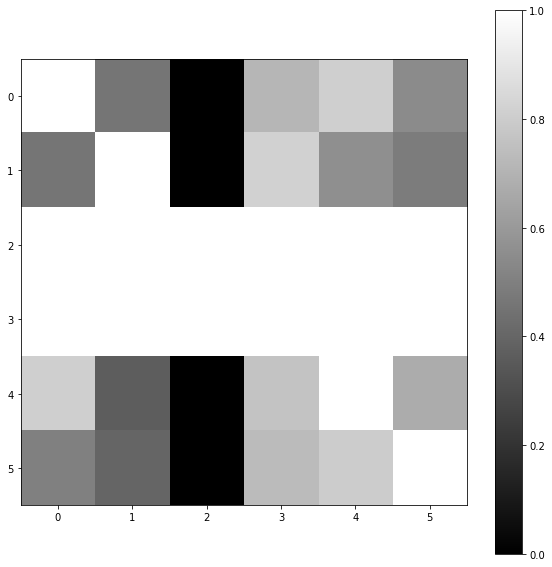}\label{fig:lenet5_ly1_cc}}
    \subfigure[LeNet-5: Layer 2]{\includegraphics[width=0.14\textwidth]{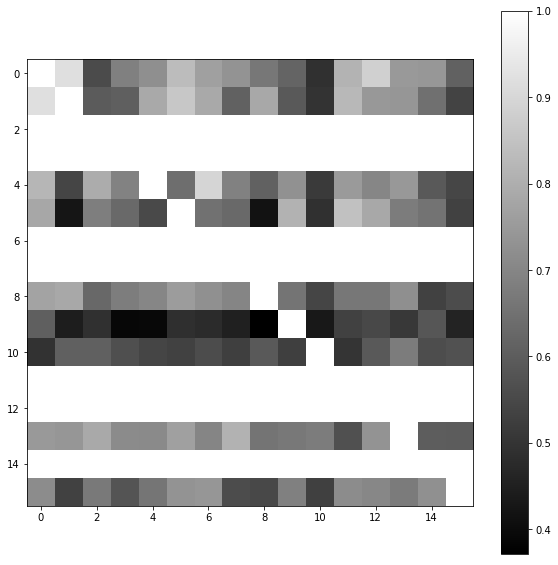}}
    
    \subfigure[LeNet-510: Layer 1]{\includegraphics[width=0.2\textwidth]{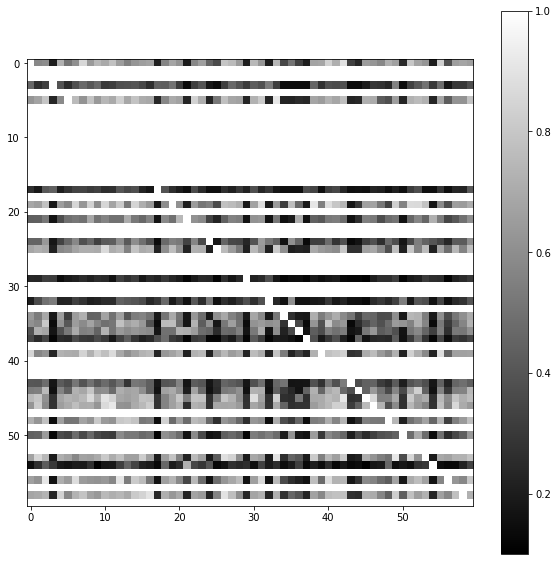}}
    \subfigure[LeNet-5x10: Layer 2]{\includegraphics[width=0.2\textwidth]{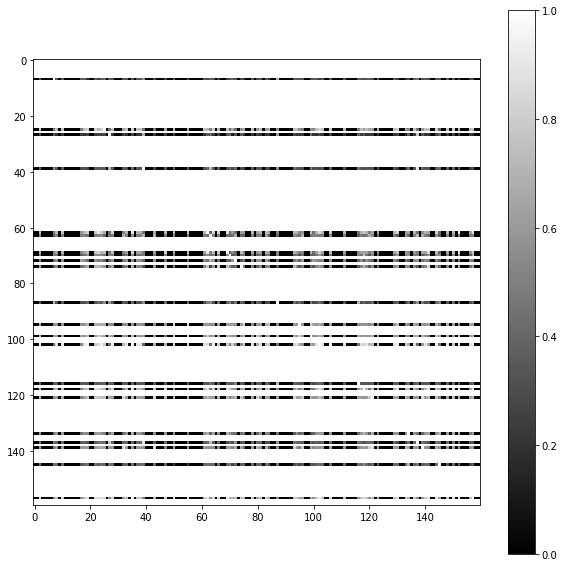}}
    \caption{Normalized cross-correlation between the impulse response of kernels in (a) layer 1 and (b) layer 2 of LeNet-5; (c) first and (d) second layer of LeNet-5x10 respectively.}
    \label{fig:lenet5_cc}
\end{figure}

Impulse response analysis is performed on LeNet-5 and LeNet-5x10. The response of kernels in the two convolutional layers are shown in Figure~\ref{fig:lenet_impresp}. Cross-correlation between the activation of a layer is computed in a one vs. all fashion and the maximum value of correlation is presented in Figure~\ref{fig:lenet5_cc}. As an example, the response of 6 kernels in the first layer of LeNet-5 produces a $6\times 6$ correlation matrix as shown in Figure~\ref{fig:lenet5_ly1_cc}.

\subsection{Discussion}
\label{subsec:imp_disc}
It is observed from Figure~\ref{fig:lenet5_cc} that the over-parametrized LeNet-5x10 has larger number of activation having high correlation in comparison with LeNet-5. This supports the observations from the channel and pixel shuffling experiments that LeNet-5x10 has higher redundancy. The highly correlated responses are produced by duplicate kernels in a layer and also due to the effect of cascaded convolutions which produce similar response as discussed in Section~\ref{subsec:imp_th}.

\section{Conclusion}
\label{sec:conc}
In this work, we have presented an approach for understanding the robustness of a CNN which is observed on account of redundancy in it. Robustness of a CNN owing to redundant kernels are identified by introducing perturbations in the input to the convolutional layers in the form of \emph{channel shuffling} or \emph{pixel shuffling}. The proposed method was experimented on LeNet5 and over-parametrized version of it, LeNet-5x10. Further, presence of redundancies were quantitatively measured using unit impulse response of the network. Cross-correlation based analysis showed that over-parametrized network has higher redundancy resulting from cascaded kernels producing similar response.

{\small
\bibliographystyle{ieee}
\bibliography{myref}
}

\end{document}